# Tensor Radiomics: Paradigm for Systematic Incorporation of Multi-Flavoured Radiomics Features

Arman Rahmim[1,2,3], Amirhosein Toosi[2], Mohammad R. Salmanpour[2], Natalia Dubljevic[3], Ian Janzen[4], Isaac Shiri[5], Ren Yuan[4], Cheryl Ho[4], Habib Zaidi[5], Calum MacAulay[4], Carlos Uribe[1,4], Fereshteh Yousefirizi[2]

[1]Department of Radiology, University of British Columbia, Vancouver, Canada
[2]Department of Integrative Oncology, BC Cancer Research Institute, Vancouver, Canada
[3]Department of Physics & Astronomy, University of British Columbia, Vancouver, Canada
[4]BC Cancer Research Institute, Vancouver, Canada
[5]Division of Nuclear Medicine and Molecular Imaging, Geneva University Hospital, Geneva, Switzerland

**Abstract:**

**Background and Objectives:** Radiomics features extract quantitative information from medical images, towards the derivation of biomarkers for clinical tasks, such as diagnosis, prognosis, or treatment response assessment. Different image discretization parameters (e.g. bin number or size), convolutional filters, segmentation perturbation, or multi-modality fusion levels can be used to generate radiomics features and ultimately signatures. Commonly, only one set of parameters is used, resulting in only one value or 'flavour' for a given radiomics feature. We propose 'tensor radiomics' (TR) where tensors of features calculated with multiple combinations of parameters (i.e. flavours) are utilized to optimize the construction of radiomics signatures.

**Methods:** We present examples of TR as applied to PET/CT, MRI, and CT imaging invoking machine learning or deep learning solutions, and reproducibility analyses: (1) TR via varying bin sizes on CT images of lung cancer and PET-CT images of head and neck cancer for overall survival (OS) prediction. A hybrid deep neural network, referred to as 'TR-Net', along with two machine-learning-based flavour fusion methods were applied. (2) TR built from different segmentation perturbations and different bin sizes for classification of late-stage lung cancer response to first-line immunotherapy using CT images. (3) TR via multi-flavour generated radiomics features in MR imaging of glioblastoma. (4) TR via multiple PET/CT fusions in head and neck cancer. By using Laplacian pyramids and wavelet transforms, flavours were created based on different fusions.

**Results:** Our results showed that TR based on varying bin sizes on CT images of lung cancer and PET-CT images of head and neck cancer improved accuracy compared to regular radiomics features for OS prediction. A TR derived from different segmentation perturbations and different bin sizes on CT images improved classification of lung cancer response to therapy. TR applied to MR imaging of glioblastoma showed improved reproducibility when compared to many single-flavour features. In head and neck cancer, TR via multiple PET/CT fusions improved overall survival prediction.

**Conclusion:** Based on our results, we conclude that the proposed TR paradigm has significant potential to improve performance in different medical imaging tasks.

## 1. Introduction

The term 'radiomics' was first introduced in 2010 by Gillies et al. [1] as "the extraction of quantitative features from radiographic images". Radiomics features capture information about tissues and lesions [2]

[3]–[5]. A collection or combination of radiomics features considered to be a 'radiomics signature' can computationally model a biological phenomenon [6]. Singular radiomics features and radiomic signatures can act as imaging biomarkers and have been shown to reflect biological characteristics of lesions and can improve a range of different clinical tasks [7]–[9]. Radiomics includes the use of single- or hybrid-imaging modalities, with the potential to identify novel imaging biomarkers for improved detection, classification, staging, prognosis, prediction and treatment planning in different cancers [9].

A radiomics feature can be generated using different parameters (e.g. pre-processing, discretized bin number or size; segmentation threshold to define region-of-interest). There have been significant efforts to establish the best values of parameters suitable for different tasks. However, determining suitable predictive features can be a difficult task. Certain features or versions of features (with specific parameters) may be non-robust to noise or can change depending on the scanner used to acquire the imaging data. This can result in a lack of reproducibility across different institutions and scanners [14]–[18] and radiomics features may be non-robust to noise or inter-center protocol and scanner variabilities [11] [12]. The vast array of feature-selection methods used in radiomics studies [19]–[21] attests to the need for careful pruning of features beforehand and the difficulty of such a task. Individual radiomics features may correlate with one another, hence providing no added predicted value for a radiomic signature, or they may challenge interpretability, such as many "deep features" extracted from layers of a neural network [3], [10]. This can result in a lack of reproducibility across radiomics trials. Although efforts have been made to standardize radiomics protocols [13] and reduce the effect of different scanners, these remain relevant issues [14] [20].

This work aims to tackle the above-mentioned limitations of radiomics analyses using a different paradigm. We propose 'tensor radiomics' (TR) in which multiple flavours (i.e. versions of the same radiomics features) are generated and considered. Generally, a radiomics feature consists of a single value for an entire 2D or 3D volume. Instead of providing a single value for each feature, we compute multiple values of the feature by varying some of the parameters in the calculation. Features with different bin sizes, perturbation of segmentations, pre-processing filters (e.g. Laplacian of Gaussian (LOG) and Wavelets), and fusion techniques can be considered as different flavours. We hypothesize that TR has the potential to overcome some of the shortcomings of models that use radiomics features enabling improved clinical task performances.

## 2. Patient Data and Methods

### 2.1. Patient data

In this paper, we explore application of the proposed TR paradigm to a range of data and applications, including CT images of lung cancer, PET-CT images of head and neck cancer, and MR imaging of glioblastoma (**Table 1**).

Table 1. Data description and the corresponding tasks

| Dataset | Modality | Task | Number of cases |
|---|---|---|---|
| **Lung** | CT | Treatment response prediction | 118 lung lesions (primary and lung metastases) from 96 patients |
| **Head and Neck** | PET/CT | 2-year progression free survival | 224 |
| **Brain** | MRI | Repeatability analysis | 17 |

### 2.2. Methods

Our original intuition to explore the TR paradigm was that features generated using multiple flavours may be more robust and reproducible (and ultimately valuable) compared to single-flavour features for outcome prediction. In TR, we build feature tensors using many flavours (**Fig. 1**) as a method towards the optimized construction of radiomics signatures. For different clinical tasks, subsequent to the generation of the radiomics tensor, one may utilize feature selection or extraction methods, as well as machine learning (ML) or deep learning (DL) methods, for optimal construction of signatures.

In what follows we further elaborate possible approaches to TR and provide example applications in PET/CT, MRI, and CT imaging. We generate multi-flavour TR signatures by varying bin sizes in PET-CT (**section 2.1.1**) and CT images (**section 2.1.1**). Different segmentation perturbations are our next approach to build different flavours of radiomics signature in CT images (**section 2.1.2**). Multi-flavour radiomics features generated by varying hyperparameters of pre-processing filters are also considered in MRI images (**section 2.1.3**). Flavours built from different fusions methods on PET-CT images are also considered towards building TR signature in this effort (**section 2.1.4**).

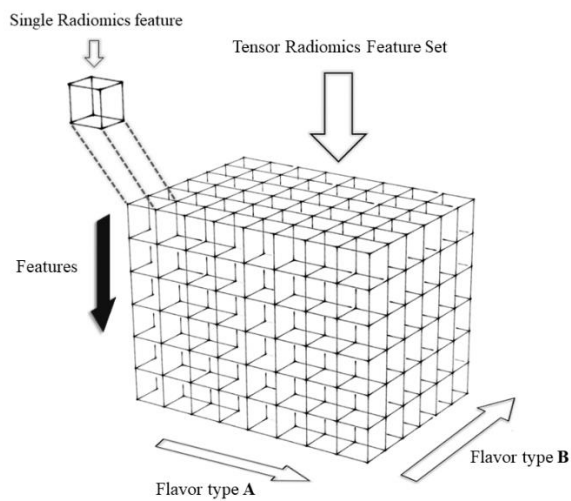

**Figure 1.** A representation of a radiomics tensor. The black arrow indicates the dimension where the different extracted radiomics features are stacked, while the flavour types (A, B, ...) encompass variations by which feature values are generated: examples include variations in discretization bin sizes, segmentations, pre-processing filters and fusion methods. Only two variants can be represented here in 3D, but even higher dimension tensors can be generated.

### 2.1.1. Discretization flavours

**Data and task:** We evaluated whether TR may enhance the prediction of outcome using PET/CT scans in patients with HNC. The 224 baseline HNC PET/CT scans were obtained from The Cancer Imaging Archive (TCIA), accompanied with tumor expert-manual segmentations. In this work, we formulated the outcome prediction task as binary classification (2-year progression free survival). TR features were extracted from PET images using 10 different bin sizes (0.1-1.0 SUV) using the *PyRadiomics* package [23] (106 features used).

The primary realization of proposed TR utilizes explainable, handcrafted radiomics features at varying discretization levels. Discretization is the grouping of the original range of pixel intensities into specific intervals or bins, necessary for the computational feasibility of certain features [22]; e.g. fixing the bin widths (BW) or fixing the number of bins (bin counts (BC)). Different 'bin flavours' calculated by different discretization strategies are shown in **Fig. 2**. An optimal discretization would be one that consistently filters out noise while retaining the integrity of important lesion features. However, little guidance is given on achieving such an optimum in the literature, even though discretization can dramatically impact the calculated feature values. In our proposed TR paradigm, a variety of bins (flavours) can be generated, and the best ones are ultimately utilized (Please consider **Fig. S1** for further details).

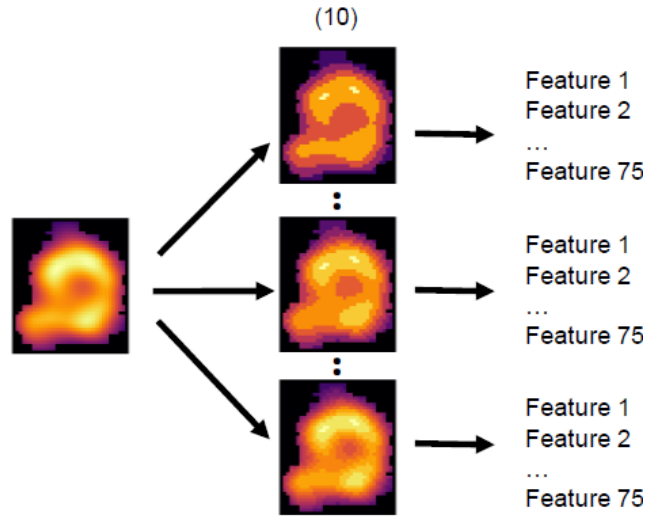

Figure 2. Radiomics features calculated on PET/CT images from head and neck data. 10 'bin flavours' calculated by using different discretization strategies to extract 75 features.

*ML-based flavour selection framework.* First, we trained two ML models, support vector machines (SVM) with radial basis function (RBF) kernel and logistic regression, on concatenated features for all possible combinations of 10 flavours (1013 different combinations with 2 or more flavours). The TR models were compared to the same ML models trained on single-flavour features.

*DL-based flavour fusion framework.* We developed a two-stage deep feed-forward neural network, named TR-Net (for tensor radiomics Network), to perform end-to end-flavour fusion (see **Fig. 3** for further details). The input to this model were all features extracted using all flavours. TR-Net consists of legs and body: each leg is a stack of multiple dense layers separately fed with the features of each flavour in its input layer. All legs are then concatenated and connected to a few more dense layers to complete the architecture of the network. The size of the final dense layer, as well as the size of other dense layers in the architecture and the number of the layers in both the legs and the head part, were among the hyperparameters that we tuned them using GRID search on the validation set.

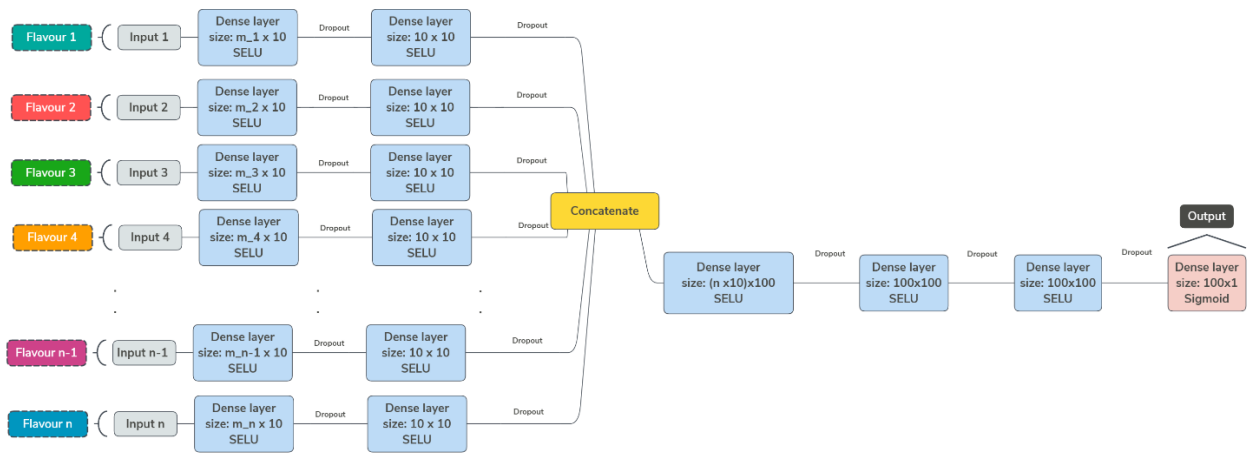

Figure 3. The architecture for our TR-Net (Tensor Radiomics Network). Features of a given flavour are input into the 'legs' of the network, which work to extract and combine the most effective features. The various flavour legs are then concatenated and put through several final layers before producing a binary prediction.

A sigmoid function in the last layer performs the binary outcome classification. The network was trained using a mean squared error (MSE) loss function. Random selection search using a nested 5-fold cross-validation was utilized for hyperparameter tuning. Our TR-Net is a modified version of a fusion network initially proposed in [24] for biometrics liveness detection. Intuitively, during training, TR-Net concurrently learns a transformation of features coming from its legs (different flavours in our case) into a common subspace (in the first stage) and performs classification on those features based on their fusion in that common subspace (in the second stage). To tackle class imbalance during training the network, SMOTE [25] was applied to the training set only, to up-sample the minority (positive) class.

Average balanced accuracy and F1 score metrics were computed using stratified 5-fold cross-validation (CV) (for both ML & DL methods). Scaled Exponential Linear Unit (SELU) activation function [26] was used in all the dense layers of the legs and the body along with dropout.

*ML-based feature selection framework.* Motivated by the improved performance of TR-Net, we studied the effect of applying feature selection (for any given flavour across different features) prior to performing flavour fusion, on performance of the classification task. A range of 5 to 25 features were selected from each of the 10 flavours using sequential forward feature selection method [27] based on the mean f1 score over a nested 5-fold cross validation setup. These selected features were then combined for all possible combinations of 10 flavours to form our TR features, and then ranked based on mean balanced accuracy and f1 score of nested 5-fold cross validation using two ML classifiers, namely an RBF kernel SVM and a Logistic Regression model.

### 2.1.2. Segmentation flavours

**Data and task**: We analyzed the viability of TR features with a Lung Volume CT dataset of late-stage Non-small Cell Lung Cancer (NSCLC) patients. This dataset is a retrospective cohort of patients with Programmed Death Ligand-1 (PDL1) scores ≥ 50% that received first-line single-agent pembrolizumab as the current Canadian clinical practice [28]–[30]. Two oncology radiologists read the baseline and 1st follow-up CT to assess treatment response (e.g. Disease Control, including: Complete/Partial Response (C/PR) and Stable Disease (SD) vs Progressive Disease (PD)) using RECIST v1.1 [31]. This CT dataset consists of 118 lung lesions (primary and lung metastases) from 96 patients, with up to 5 axial adjacent slices per lesion selected to provide a more supplementary descriptive CT image dataset (n=558). The ML task for this dataset was to predict patient response to pembrolizumab (i.e., Disease Control) using 2D-only TR features extracted from baseline axial CT lesion segmentations.

We also explored TR based on segmentation flavours. Inspired by Zwanenberg et al. [32] where image perturbations to assess feature robustness were used, we incorporated combinations of translation, area adaptation, and contour randomization of the images and masks to extract a number of perturbed images per axial slice of available segmentations. This TR segmentation flavor technique was employed to leverage an established test-retest method in an effort to establish robust and stable features. For each of these perturbed images, radiomic features were extracted with *PyRadiomics* and used to build our radiomics tensor. Our TR feature set was used to train a ML model to execute this binary patient response task for measurement of disease control (PR+SD vs PD). We leveraged Zwanenberg et al.'s methodology, primarily the "volume adaptation" (V) method, for our lung lesion ROIs to extract static bin size feature flavours from perturbed 2D lesion segmentations. This V method describes a process wherein the ground truth mask is either dilated or eroded using a disc-structuring element. An arbitrary and equal number of dilations and erosions are performed to capture extra parenchymal tissue either (dilation) or tissue characteristics from the core of the lesion only (erosion). To further supplement the protocol from section 2.1.1 (bin size flavours), each vanilla feature in the 2D CT feature set was recalculated using increasing bin width sizes for extraction of TR features without image perturbation. We compared the validity of both schemes of TR features, bin size changes and image perturbations respectively, with a standard rudimentary ML model, the Linear Discriminant Analysis (LDA), for design simplicity. These models were trained on TR features chosen by Sequential Forward Feature Selection (SFS). An LDA trained on 7 vanilla radiomics features

(i.e., non-perturbed images and a static bin width) selected by SFS was chosen as the primary analysis method for comparison (5-fold nested CV and patient agnostic training-test splitting to avoid biasing the results; SMOTE resampling for class imbalance). A statistical power analysis revealed that this late-stage cancer dataset sample size (N=96), achieves 80% power for detecting a medium effect (ES=0.61) at a significance level α=0.05 for Welch's t-test. Therefore, given our positive event rate (26:70 positive:negative patients), a 5-fold CV was selected provided that a medium effect size is maintained for each fold given the approximate number of patients in the test set.

We compared a series of TR feature flavours against a baseline LDA model trained on unperturbed images. We refer to this baseline as having "vanilla" feature flavours. We created 3 other TR feature flavour sets from images to test against this vanilla model: 1) recalculated features using increasing bin widths; 2) segmentation volume adaptation via dilating and eroding the ROI mask; and lastly 3) a combination of image translation ("T"), segmentation volume adaptation (V), and segmentation contour randomization ("C") to produce the "TVC" TR feature flavour set.

### 2.1.3. Filter (pre-processing) flavours

**Data and task**: We analyzed MR images of 17 glioblastoma (GBM) patients including T1- and T2-weighted images (performed within the same imaging unit on two consecutive days; full affine registration of test to retest images with 12 degrees of freedom using mutual information cost functions)[33]. N4 bias correction was performed on raw MR images). Our task in this experiment is assessing the repeatability of MR imaging radiomics features.

In this approach, tensors of radiomic features were generated after applying pre-processing filters as different flavours such as wavelets (WL, all possible combinations of applying either a high- or a low pass filter in each of the three dimensions, including HHH, HHL, HLH, HLL, LHH, LHL, LLH, and LLL), LOG (with different sigma values of 0.5 to 5 with steps 0.5), Exponential, Gradient, Logarithm, Square and Square Root scales. In addition to filter flavours, we also used discretization flavours (e.g. section 2.1) of fixed bin width and fixed bin count (16, 32, 64, 128, and 256). Different features, namely first-order (FO) and textural features (GLCM, GLRLM, GLDM, GLSZM, NGTDM) were then extracted. Principal component analysis (PCA) was performed on different flavours to generate new features. One component per feature was chosen (converting multiple preprocess features to one feature). We performed this task regarding the improvement of repeatability of radiomics features across different preprocessing steps. Intraclass correlation coefficients (ICC) were calculated to assess the repeatability of features (comparing performances of conventional single-flavour features vs. newly-generated features). ICC value classified to ICC<50%, 50%<Value<75%, 75%<Value<90% and 90%<Value<100% for low, medium, high, excellent repeatable features.

### 2.1.4. Fusion flavours

**Data and task:** To evaluate fusion flavours, we used the same data and task as in Sec. 2.2.1 (head and neck cancer PET-CT binary outcome prediction).

Fusion radiomics [34], an emerging area of investigation, has up to now meant fusing images in different ways and selecting the optimal one. In the proposed TR, we alter the paradigm and include various multi-modality fusions within the radiomics tensor for a given task, followed by subsequent model construction. We employed 15 image-level fusion techniques to combine PET-CT information (see Table S1). Subsequently, 211 features were extracted from each region of interest in PET-only, CT-only, and 15 fused PET-CT images through the SERA radiomics package [35]. A range of optimal algorithms was pre-selected amongst various families of learner algorithms.

We considered 2 approaches for the prediction task: (i) We separately applied radiomics features extracted from each of the PET-only, CT-only, and 15 fused PET-CT images to 3 classifiers, namely logistic regression, random forest, and multilayer perceptron (MLP) classifiers. We employed ensemble voting for each classifier: specifically, we used 9 different estimators for each classifier (i.e. with different optimized parameters from 9 different runs/initializations/grid-searches). (ii) For the proposed TR paradigm, we incorporated 211 radiomics features with all 17 flavours, removing features with correlations over 95% to avoid redundant data. Subsequently, Polynomial Feature Transforms were applied to fuse the selected relevant flavours. Polynomial features transform is a simple and common method to specify feature's high-order and interaction terms. It generates a new feature matrix containing all polynomial combinations of the features with degrees less than or equal to the specified degree. After normalizing the fused features, we applied ANOVA F-value feature selection algorithm [36] to select the most relevant fusion-TR features. Finally, we applied these selected fusion-TR features to the 3 above-mentioned classifiers to predict survival outcome. In this work, we performed nested 5-fold cross validation. In each round, we divided the dataset into a training part (four-fold) and testing part (one-fold). In the training process, we further divided the training dataset into 2 sub-parts, with 80% of datapoints for training the model and remaining 20% for model selection. Mean accuracy in training validation was used to select the best model. Mean accuracy in nested testing was reported to validate the best model (for more information please consider Fig. S2).

## 3. Results

### 3.1. Results of using discretization flavours

We first assessed the effectiveness of combining features from different combinations of the flavours (bin size in this part of our study) in terms of 2 classification score metrics: balanced accuracy, and f1 score (we also studied area under ROC curve, with similar trends; not shown). **Fig. 4** illustrates the enhanced metric values using 5 different flavour combinations taken from the top 20 combinations out of all possible ones. The rationale behind this comparison is to examine whether combining "all" features of multiple flavors improve the classification, without performing feature selection, and whether the combinations that improve the classification performance, show some meaningful combinations; e.g. evaluating if the combination of the smallest bin size with the largest bin size can improve the performance. To this end, we examined all possible combinations and reported a few of the better performers with respect to single bin size features.

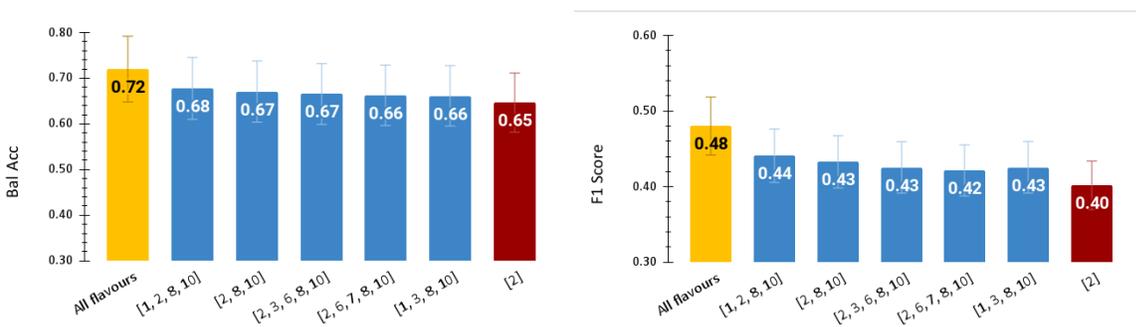

Figure 4. Balanced accuracy (Left) and f1 score values (Right) when using conventional radiomics (single-flavour) using ML pipeline (red), 5 different combinations of TR flavours using ML pipelines (blue), and all 10 TR-flavours of features via DL TR-Net pipeline (yellow). All features were used (no features selection methods were applied prior).

Our results show that TR features made from multiple flavours with respect to regular radiomics features improved performances from (red) conventional radiomics to (blue) TR in ML pipelines. Further improvements were obtained when utilizing (yellow) our DL TR-Net pipeline in an end-to-end fashion. We also studied the effect of feature selection prior to ML methods applied to conventional vs. TR models, with similar trends (see **Fig. 5**). We applied a corrected t-test on the results but did not detect significant outperformance relative to the baseline.

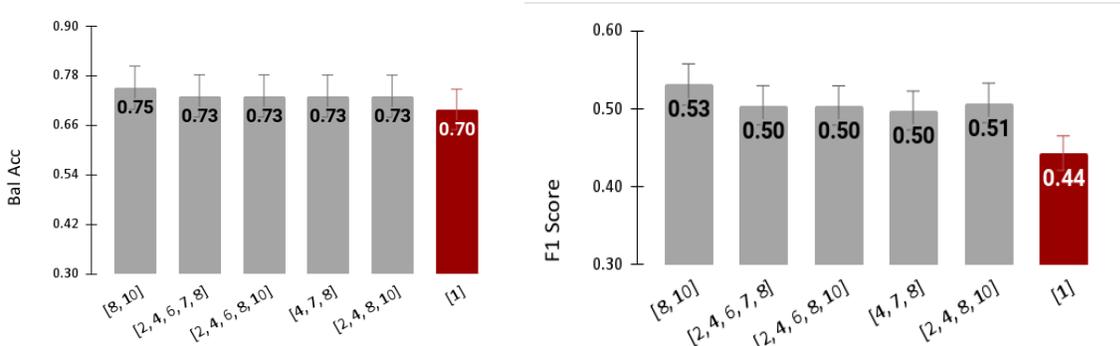

Figure 5. Balanced accuracy (Left) and f1 score values when using feature selection (Right), prior to applying conventional radiomics (single-flavour) (red) vs. TR (multi-flavours) (gray).

## 3.2. Results of using segmentation flavours

ROC and Precision-Recall analysis on vanilla feature flavor, different bin size, the combination of image translation ("T"), segmentation volume adaptation ("V") and contour randomization ("C") were conducted as seen in **Fig. 6**.

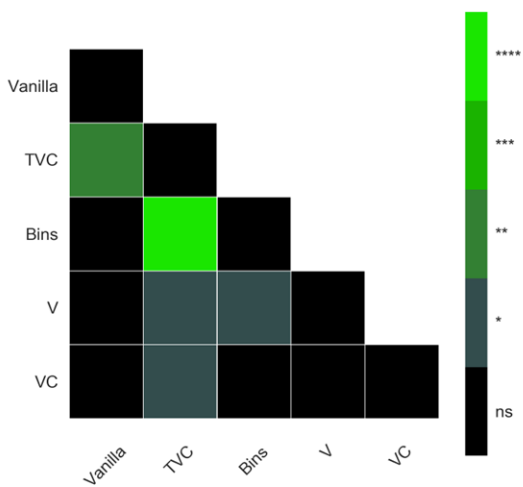

Figure 6. Significance matrix of 5 discrete LDA models in predicting treatment response to single agent pembrolizumab in late-stage NSCLC patients using radiomic features. Each model was trained with a unique feature set (image and segmentation flavours). McNemar's test was applied to test the significance between model performances shown in the matrix (color) to the right.

Table 2: The performance of the LDA model on the test folds. ROC AUC = Receiver Operator Characteristic Area Under the Curve; mAP = mean Average Precision.

| Flavour | Balanced Accuracy | F1 Score | ROC AUC | mAP |
|---|---|---|---|---|
| Vanilla | 0.67 ± 0.08 | 0.79 ± 0.08 | 0.73 ± 0.12 | 0.88 ± 0.09 |
| TVC | 0.77 ± 0.06 | 0.85 ± 0.07 | 0.83 ± 0.09 | 0.92 ± 0.08 |
| Bins 1-to-100 | 0.73 ± 0.08 | 0.82 ± 0.09 | 0.78 ± 0.06 | 0.92 ± 0.04 |
| V | 0.68 ± 0.08 | 0.79 ± 0.09 | 0.65 ± 0.14 | 0.84 ± 0.10 |
| VC | 0.69 ± 0.09 | 0.87 ± 0.02 | 0.82 ± 0.06 | 0.93 ± 0.03 |

To assess the effect of our proposed segmentation TR flavours on the outcome prediction task, Group-k (k=5) fold CV was performed to maintain patient agnostic training and test folds. Reported metrics and uncertainties were based on the average scores of area under the ROC curve (ROC AUC), and mean average precision (mAP) and F-1 score using this CV technique. The results in terms of these classification metrics were compared to the single-flavoured (vanilla) features. Furthermore, the performance of the LDA model on the test folds by applying feature selection prior to model training is illustrated in Table 2 as well as Fig. 6 and 7.

Overall, the trend indicates a strong model performance with the addition of TR feature flavours. We observe boosts to established metrics such as ROC AUC and PR AUC over the vanilla feature model as evidenced by Table 2 (and **Fig. 7**). TVC perturbation maintained the largest F-1 score of 0.852±0.09 in comparison to the vanilla model's 0.793±0.08. The weakest model was the solo volume adaption (V) feature flavour set in nearly all metrics. However, looking at **Fig. 7**, one can observe a non-small model improvement in the image perturbation feature sets (TVC and V) at low false positive rates. All in all, our results suggest the TR paradigm to enable improved performance relative to conventional radiomics.

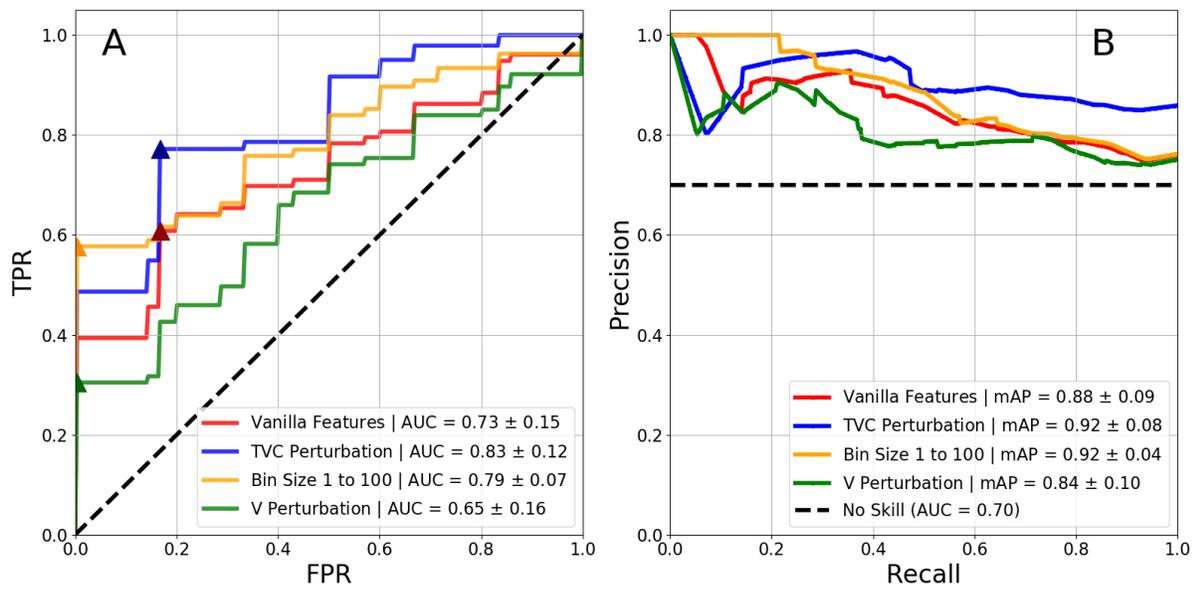

Figure 7. Model performance plots for 4 discrete LDA models. A: ROC curve plot highlighting model performance with the addition of feature flavour segmentations. TVC Perturbations achieved the highest ROC AUC = 0.83±0.12, SN: 77.9%, SP: 83.2% at optimal threshold (triangle marker). B: Precision-Recall curve to elucidate model performance for this class imbalanced dataset. Both plots indicate a strong boost to performance of the model by implementing TR feature methods.

### 3.3. Results of using filter (pre-processing) flavours

As seen in **Fig. 8**, LOG results in the highest number of repeatable features. In LOG preprocessing with sigma 0.5 and 1 there were features with low repeatability. However, in TR there were no poorly-repeatable features, and only 2 features showed medium repeatability, with the rest showing high and excellent repeatability. In wavelet features, different settings resulted in different numbers of non-repeatable and repeatable features; LLL results in the highest number of repeatable features. TR of wavelets showed 3 low, 11 medium, 45 high and 34 excellent repeatable features. Overall, TR increased radiomics-feature repeatability in all features and only two features Kurtosis from first order and busyness from NGTDM had low repeatability (ICC<50%). In TR, features including Complexity and Coarseness from NGTDM, CP from GLCM and Skewness from first-order had medium repeatability (50<ICC<75%).

Twenty-four features showed high repeatability (75<ICC<90%) and 61 from 93 features showed excellent (ICC>90%) repeatability.

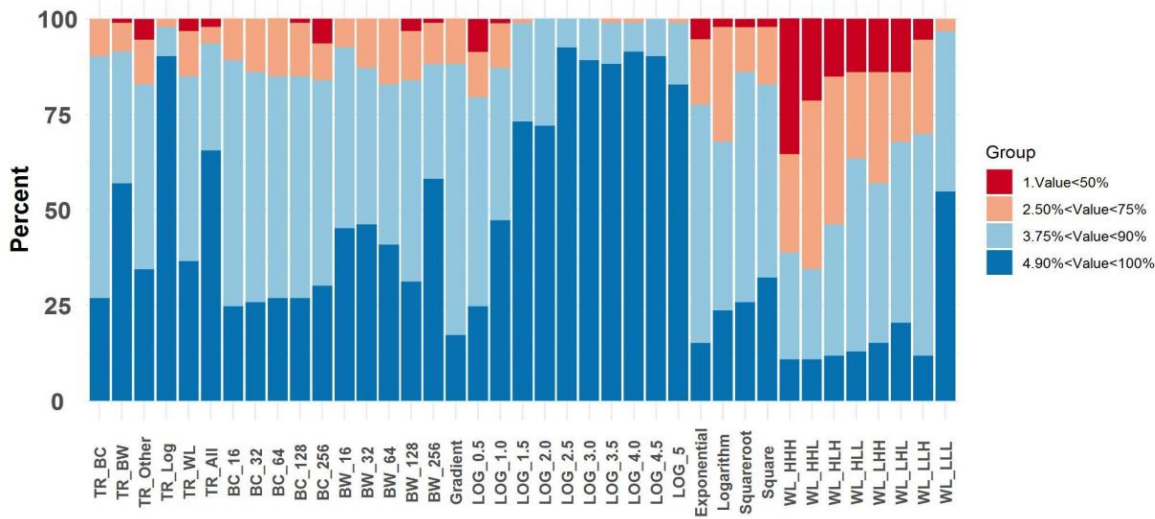

Figure 8. The test-retest repeatability of radiomics features in MR imaging of GBM (ICC used as metric). Proposed TR-generated features (via PCA) are compared to conventional individual flavours of these features.

### 3.4. Results of using fusion flavours

For our fusion-based TR framework, we included 211 radiomics features. After feature fusion, we selected the relevant fusion-TR features, applying them to the mentioned ensemble classifiers. The highest training validation performance of 73.6 ± 2.5 % was obtained for fusion-TR via MLP, and nested testing of 71.8 ± 4.8 % confirmed this finding as shown in **Fig. 9**. For conventional non-TR scheme, the highest training validation performance of 69.7% ± 5.1% was obtained for Laplacian Pyramid (LP) + Sparse Representation (SR) linked with random forest, with nested testing performance of 66.6% ± 3.9%. Paired t-test indicated significantly improved ($p<0.05$) performance for the proposed TR framework compared to best non-TR performance.

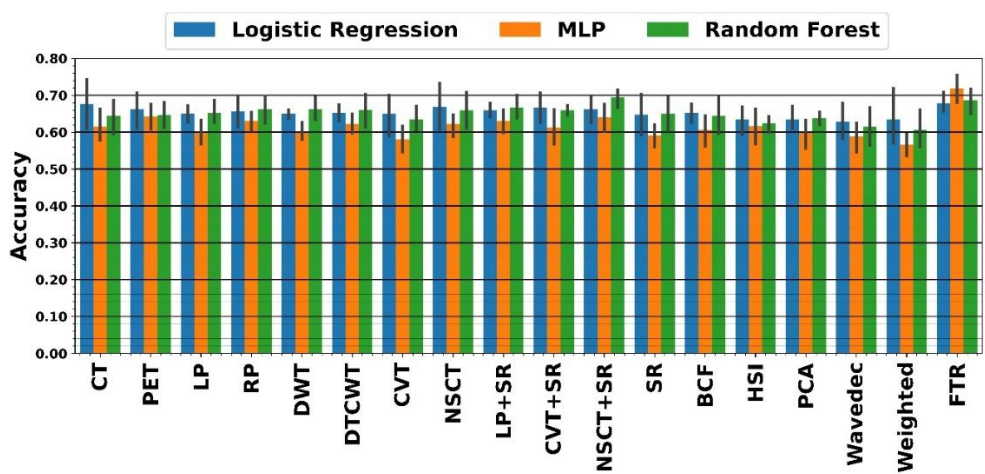

Figure 9. Bar plots of Mean & STD for prediction of outcome for nested testing performance. X axis: conventional radiomics (PET-only, CT-only, 15 fused PET/CT images), as well as proposed fusion-TR (FTR), using 3 different classifiers.

## 4. Discussion

Radiomics features can improve clinical task performances by capturing important pathophysiological information [7]. In routine practice, radiomics features are included using a fixed set of parameters. In this study we generated TR features by varying multiple parameters such as bin sizes, exploring different segmentation perturbations, varying hyperparameters of pre-processing filters and testing different fusion techniques. We explored the tasks of survival prediction outcomes (**sections 3.1, 3.2 and 3.4**) as well as test-retest repeatability (**section 3.3**). Our proposed TR approach allows the learning algorithms to consider different aspects of imaging features, moving beyond conventional paradigms.

Applying bin size variations to create different flavours of TR features for the HNC study described in section 3.1 showed its effectiveness in performance enhancement of the outcome prediction task. Compared to the radiomics features extracted in a regular manner, combining the different bin size radiomics features flavours presented more discriminative power for the classification problem. However not all features from all flavours combined necessarily lead into a more discriminative set of information. Employing flavour combination selection and TR-Net for end-to-end flavour fusion back up this statement (Fig. 4). By using SELU activation functions along with dropout regularization in every layer, TR-Net can learn a transformation of features from all flavours to a common latent space in a way that it keeps the contribution of informative features coming from different flavours while decreasing their weaknesses through suppressing the redundant features. Furthermore, motivated by the promising results obtained by TR-Net compared to non-DL results achieved by flavour combinations, we investigated the idea of performing feature selection per each flavour to make our TR feature set less redundant and more discriminative. Prior to applying ML and DL pipelines to their combinations, it is possible to use correlation between the same features from different flavors to remove the highly correlated ones. This work is basically trying to represent proof of the concept of Tensor Radiomics and we believe that this is an important refinement that needs to be explored in future studies.

For the study of segmentation flavors in **section 3.2**, a majority of LDA model performances saw an overall boost with the addition of TR feature flavours for segmentation perturbation as TR flavours. Notably, the more aggressive image perturbation technique ("TVC") achieved the highest ROC AUC score and second highest PR AUC as seen in **Table 2**. The strong performance of the differing bin size TR feature set over the vanilla features also supports the findings of section 3.1 vis-a-vis radiomic feature classification power improvement via changing bin size when calculating radiomic features. Continuing to look at **Fig. 7** the TVC and bin size methods also showed a marked improvement over the vanilla features at low false positive rates, indicating a higher sensitivity at these thresholds. In the context of this task, identifying patients who are less likely to respond to pembrolizumab with a high specificity while maintaining a sensible sensitivity is a crucial finding in patient survival. The models that are able to identify PD patients (i.e. non-responders) allow clinical decision makers to adjust from 1st-line immunotherapy with pembrolizumab ("pembro") to combination therapy pembro and platinum-doublet chemotherapy in an effort to more aggressively combat late-stage tumor growth. The findings of the study in section 3.2 indicate the power of TR based features in identifying clinicopathological biomarkers that can elucidate medical utility for this homogenous patient population.

In **section 3.3**, we considered TR based on different hyper-parameters of pre-processing filters (LOG and wavelet) increased radiomics-feature repeatability. TR made by different hyper-parameters of LOG preprocessing filter did not show any poorly-repeatable features, and only 2 features had medium repeatability. On the other hand, TR based on different flavours of wavelet filter, had some non-repeatable and repeatable features. TR based on fusion flavours also showed significant improvement over the best non-TR performer (**section 3.4**). The fusion flavor idea can be further extended to additionally include deep features from different image-fusions via deep neural networks (e.g. autoencoders), adding those to existing TR features to study potentials to further enhance performance.

Beside the effective methods employed in different studies carried out throughout the previous sections of this work, in the frame of TR, there were many other studies that were tested but not all were fruitful. However many valuable lessons were learnt that are indeed worth mentioning here. As a case in point, for bin size flavours study (**section 3.1**), we tested many other implementations involving binning flavours, but not all were led to meaningful improvements. Among them are the combination of flavours via PCA to create new, blended features. Multiple ML pipelines consisting of an unsupervised feature selection/dimensionality reduction method (univariate feature selection, PCA with a linear kernel, PCA with a radial basis function kernel) and an ML classifier (logistic regression, random forest, support vector machine, k nearest neighbors) were developed which showed no consistent increase or decrease in terms of classification scores when compared to a baseline that used only one flavour.

This approach was repeated using LDA to combine flavours with similarly inconclusive results. These results compared to that of the reported ones in **section 3.1** may indicate that a more nuanced approach is necessary to effectively include new bin flavours. Potential ideas include switching out PCA for a more sophisticated method of flavour combination, such as an autoencoder. Adding more flavour types other than bin flavours could also help to make more robust TR features. In this case LDA, PCA or any other feature extraction/dimensionality reduction method could perform an effective feature/flavour combination and as a result, a better prediction performance. These findings could back up the idea that employing domain knowledge in picking the best flavour types per each study case is crucial. In other words, beside finding the best method to combine multiple flavours, finding the most informative and relevant flavour variations for the task at hand is an important factor that needs to be taken carefully.

TR is important in the context of hand-crafted radiomics because previously published studies commonly choose some parameters a priori without necessarily reliable justification. Areas for future exploration include establishing the effectiveness of TR methods on other larger datasets, improved methods (e.g. use of methods for the analysis of longitudinal data, imagining different flavours as different time-points in feature-space), and exploring new flavours (e.g. deep features as mentioned earlier). The TR paradigm enables revisiting past efforts and re-studying them in different light. For instance, subregional intratumor radiomics were defined based on individual- and population-level clustering [39]: combining feature flavours generated from different tumor partitions can result in a new form of TR. In addition, shell features reflecting the tumor microenvironment can be extracted from different sizes of peritumoral regions [40] and their combinations can also be explored in the proposed context of TR.

## 5. Conclusion

For radiomics analyses, fixed parameter values are commonly used to generate feature values (e.g. discretization (bin number or size), pre-processing filtering, segmentation, or multi-modality fusion). At best, results generated via different flavours are compared to one another. Our proposed paradigm is to move beyond this, to use radiomics tensors of features calculated with multiple combinations of flavours. We applied this paradigm to different modalities, tasks and algorithms (ML and DL). Our results, from different studies and modalities, revealed that TR has the potential to enable improved task performances.

**Acknowledgement**

This work was in part supported by the Natural Sciences and Engineering Research Council of Canada (NSERC) Discovery Grant RGPIN-2019-06467, the Canadian Institutes of Health Research (CIHR) Project Grants PJT-162216 and PJT-173231, the Swiss National Science Foundation Grant SNRF 320030_176052, and the BC Cancer Foundation. We would also like to thank Dr. Barb Melosky, Dr. Stephen Lam and Dr. Monty Martin for their assistance with the lung CT volume dataset discussed in section 2.2.

**Authors Contributions**
A. Rahmim proposed the key paradigm, supervised and verified implementations and results. A. Toosi, M. R. Salmanpour, N. Dubljevic, I. Janzen and I. Shiri developed the theory and performed the techniques and extracting the tensor radiomics from different perspectives. R. Yuan and C. Ho provide the CT volume data. H. Zaidi, C. MacAulay, C. Uribe and F. Yousefirizi supervised and discussed the findings of the different implementations of the tensor radiomics idea. All authors discussed the results and contributed to the final manuscript.

**Declaration of Interests**
Nothing declared.

**Data Sharing Statement**
Our models and related codes will be available here: https://github.com/qurit/Tensor_Radiomics.

**SUPPLEMENTAL MATERIAL**

# S.1. Fusion techniques

Multi-scale transform (MST) theories are popularly deployed in various image fusion scenarios such as classical MST-based fusion methods including pyramid-based methods like Laplacian pyramid (LP) [36], and ratio of low-pass pyramid (RP) [37], wavelet-based methods like discrete wavelet transform (DWT) [38] and dual-tree complex wavelet transform (DTCWT) [39], and multi-scale geometric analysis (MGA)-based methods like curvelet transform (CVT) [40] and nonsubsampled contourlet transform (NSCT) [41]. Generally, MST-based fusion methods consist of three steps [42]. First, decompose the source images into a multi-scale transform domain. Then, merge the transformed coefficients with a given fusion rule. Finally, reconstruct the fused image by performing the corresponding inverse transform over the merged coefficients. Sparse representation (SR) addresses the signals' natural sparsity, which is in accord with the physiological characteristics of human visual system [43]. In SR-based image processing methods, the sparse coding technique is often performed on local image patches for the sake of algorithm stability and efficiency [44]. Yang and Li [45] first introduced SR into image fusion. The sliding window technique is adopted in their method to make the fusion process more robust to noise and misregistration. In [45], the sparse coefficient vector is used as the activity level measurement. NSCT + SR, DTCWT + SR, and CVT + SR are three fusion methods that mixture of MST-based and SR-based methods. In general, the low-pass MST bands are merged with an SR-based fusion approach while the high-pass MST bands are fused using the conventional ''max absolute'' rule with a local window-based consistency verification scheme [38], [46]. CBF fuses source images by weighted average using the weights computed from the detail images that are extracted from the source images using CBF. Here, the weights are computed by measuring the strength of details in a detail image obtained by subtracting CBF output from original image[47]. Principal Component Analysis (PCA), Hue, Saturation, and Intensity (HSI), Wavelet and Weighted are other fusion methods that were used in this work [48], [49].

Table S1. List of fusions techniques explored in Sec. 3.4

| Fusion techniques | Sparse Representation (SR) | Bilateral Cross Filter (BCF) | Hue, Saturation and Intensity Fusion | Weighted Fusion |
|---|---|---|---|---|
| | Laplacian Pyramid (LP) | Principal Component Analysis (PCA) | Ratio of Low-pass Pyramid (RP) | |
| Wavelet-family fusion techniques | Discrete Wavelet Transform (DWT) | Curvelet Transform (CVT) | Dual-tree Complex Wavelet Transform (DTCWT) | Nonsubsampled Contourlet Transform (NSCT) |
| | NSCT + SR | DTCWT + SR | CVT + SR | Wavelet Fusion |

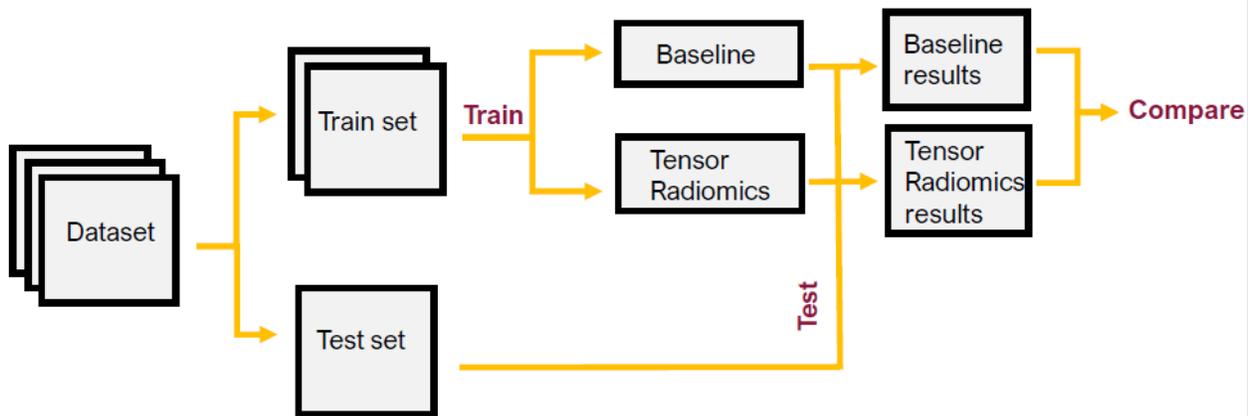

Figure S1. The Framework for TR analysis with bin size flavours

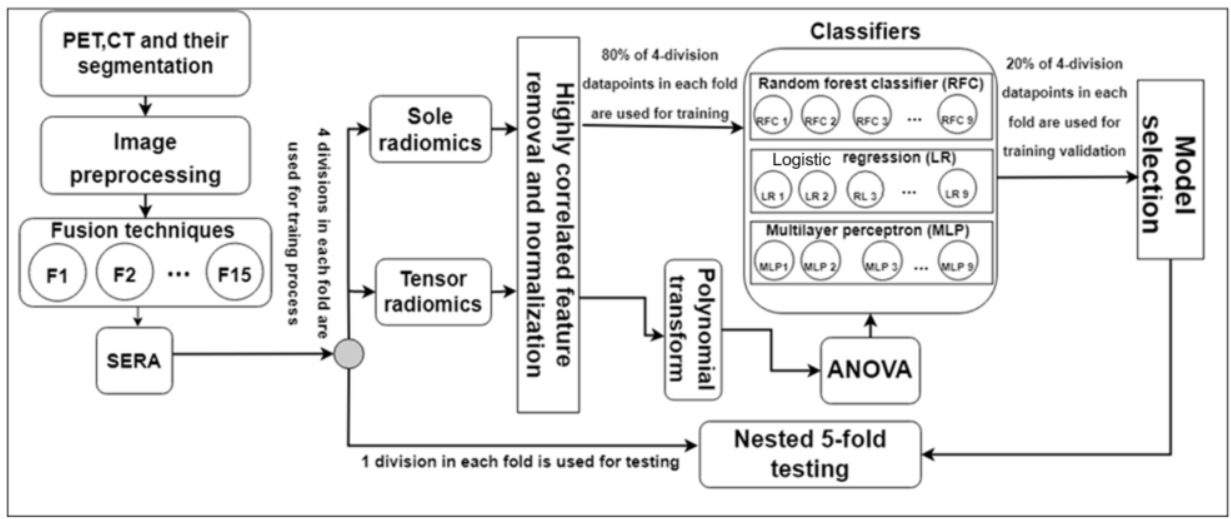

Figure S2. The suggested framework for TR with fusion

**REFERENCES**

[1] R. J. Gillies, A. R. Anderson, R. A. Gatenby, and D. L. Morse, "The biology underlying molecular imaging in oncology: from genome to anatome and back again," *Clin. Radiol.*, vol. 65, no. 7, pp. 517–521, Jul. 2010.

[2] J. E. van Timmeren, D. Cester, S. Tanadini-Lang, H. Alkadhi, and B. Baessler, "Radiomics in medical imaging—'how-to' guide and critical reflection," *Insights Imaging*, vol. 11, no. 1, pp. 1–16, Aug. 2020.

[3] F. Orlhac, C. Nioche, I. Klyuzhin, A. Rahmim, and I. Buvat, "Radiomics in PET Imaging:: A Practical Guide for Newcomers," *PET Clin.*, vol. 16, no. 4, pp. 597–612, Oct. 2021.

[4] B. Koçak, E. Ş. Durmaz, E. Ateş, and Ö. Kılıçkesmez, "Radiomics with artificial intelligence: a practical guide for beginners," *Diagn. Interv. Radiol.*, vol. 25, no. 6, pp. 485–495, Nov. 2019.

[5] A. Zwanenburg *et al.*, "The Image Biomarker Standardization Initiative: Standardized Quantitative Radiomics for High-Throughput Image-based Phenotyping," *Radiology*, vol. 295, no. 2, pp. 328–338, May 2020.

[6] Y. Huang *et al.*, "Radiomics Signature: A Potential Biomarker for the Prediction of Disease-Free Survival in Early-Stage (I or II) Non—Small Cell Lung Cancer," *Radiology*, vol. 281, no. 3, pp. 947–957, Dec. 2016.
[7] M. R. Tomaszewski and R. J. Gillies, "The Biological Meaning of Radiomic Features," *Radiology*, vol. 299, no. 2, p. E256, May 2021.
[8] J. Guiot *et al.*, "A review in radiomics: Making personalized medicine a reality via routine imaging," *Med. Res. Rev.*, vol. 42, no. 1, pp. 426–440, Jan. 2022.
[9] F. Yousefirizi, Pierre Decazes, A. Amyar, S. Ruan, B. Saboury, and A. Rahmim, "AI-Based Detection, Classification and Prediction/Prognosis in Medical Imaging:: Towards Radiophenomics," *PET Clin.*, vol. 17, no. 1, pp. 183–212, Jan. 2022.
[10] R. Da-Ano, D. Visvikis, and M. Hatt, "Harmonization strategies for multicenter radiomics investigations," *Phys. Med. Biol.*, vol. 65, no. 24, p. 24TR02, Dec. 2020.
[11] A. Zwanenburg, "Radiomics in nuclear medicine: robustness, reproducibility, standardization, and how to avoid data analysis traps and replication crisis," *Eur. J. Nucl. Med. Mol. Imaging*, vol. 46, no. 13, pp. 2638–2655, Dec. 2019.
[12] S. Ha, H. Choi, J. C. Paeng, and G. J. Cheon, "Radiomics in Oncological PET/CT: a Methodological Overview," *Nucl. Med. Mol. Imaging*, vol. 53, no. 1, pp. 14–29, Feb. 2019.
[13] R. Li, L. Xing, S. Napel, and D. L. Rubin, *Radiomics and Radiogenomics: Technical Basis and Clinical Applications*. CRC Press, 2019.
[14] G. J. R. Cook, G. Azad, K. Owczarczyk, M. Siddique, and V. Goh, "Challenges and Promises of PET Radiomics," *Int. J. Radiat. Oncol. Biol. Phys.*, vol. 102, no. 4, pp. 1083–1089, Nov. 2018.
[15] P. E. Galavis, C. Hollensen, N. Jallow, B. Paliwal, and R. Jeraj, "Variability of textural features in FDG PET images due to different acquisition modes and reconstruction parameters," *Acta Oncol.*, vol. 49, no. 7, pp. 1012–1016, Oct. 2010.
[16] I. Shiri, A. Rahmim, P. Ghaffarian, P. Geramifar, H. Abdollahi, and A. Bitarafan-Rajabi, "The impact of image reconstruction settings on 18F-FDG PET radiomic features: multi-scanner phantom and patient studies," *Eur. Radiol.*, vol. 27, no. 11, pp. 4498–4509, Nov. 2017.
[17] Y. Suter *et al.*, "Radiomics for glioblastoma survival analysis in pre-operative MRI: exploring feature robustness, class boundaries, and machine learning techniques," *Cancer Imaging*, vol. 20, no. 1, p. 55, Aug. 2020.
[18] P. Yin *et al.*, "Comparison of radiomics machine-learning classifiers and feature selection for differentiation of sacral chordoma and sacral giant cell tumour based on 3D computed tomography features," *European Radiology*, vol. 29, no. 4. pp. 1841–1847, 2019.
[19] M. E. Mayerhoefer *et al.*, "Introduction to Radiomics," *J. Nucl. Med.*, vol. 61, no. 4, pp. 488–495, Apr. 2020.
[20] I. S. Klyuzhin, Y. Xu, A. Ortiz, J. M. L. Ferres, G. Hamarneh, and A. Rahmim, "Testing the ability of convolutional neural networks to learn radiomic features," *bioRxiv*, medRxiv, 23-Sep-2020.
[21] M. Hatt, M. Vallieres, D. Visvikis, and A. Zwanenburg, "IBSI: an international community radiomics standardization initiative," *J. Nucl. Med.*, vol. 59, no. supplement 1, pp. 287–287, May 2018.
[22] J. J. M. van Griethuysen *et al.*, "Computational Radiomics System to Decode the Radiographic Phenotype," *Cancer Res.*, vol. 77, no. 21, pp. e104–e107, Nov. 2017.
[23] S. S. F. Yip and H. J. W. Aerts, "Applications and limitations of radiomics," *Phys. Med. Biol.*, vol. 61, no. 13, p. R150, Jun. 2016.
[24] A. Toosi, A. Bottino, S. Cumani, P. Negri, and P. L. Sottile, "Feature Fusion for Fingerprint Liveness Detection: a Comparative Study," *IEEE Access*, vol. 5, pp. 23695–23709, undefined 2017.
[25] N. V. Chawla, K. W. Bowyer, L. O. Hall, and W. P. Kegelmeyer, "SMOTE: Synthetic Minority Over-sampling Technique," *J. Artif. Intell. Res.*, vol. 16, pp. 321–357, Jun. 2002.
[26] G. Klambauer, T. Unterthiner, A. Mayr, and S. Hochreiter, "Self-Normalizing Neural Networks," *Adv. Neural Inf. Process. Syst.*, vol. 30, 2017.

[27] D. W. Aha and R. L. Bankert, "A Comparative Evaluation of Sequential Feature Selection Algorithms," in *Learning from Data: Artificial Intelligence and Statistics V*, D. Fisher and H.-J. Lenz, Eds. New York, NY: Springer New York, 1996, pp. 199–206.
[28] M. Reck *et al.*, "Pembrolizumab versus Chemotherapy for PD-L1–Positive Non–Small-Cell Lung Cancer," *N. Engl. J. Med.*, vol. 375, no. 19, pp. 1823–1833, Nov. 2016.
[29] L. Gandhi *et al.*, "Pembrolizumab plus Chemotherapy in Metastatic Non–Small-Cell Lung Cancer," *N. Engl. J. Med.*, vol. 378, no. 22, pp. 2078–2092, May 2018.
[30] L. Paz-Ares *et al.*, "Pembrolizumab plus Chemotherapy for Squamous Non–Small-Cell Lung Cancer," *N. Engl. J. Med.*, vol. 379, no. 21, pp. 2040–2051, Nov. 2018.
[31] E. A. Eisenhauer *et al.*, "New response evaluation criteria in solid tumours: revised RECIST guideline (version 1.1)," *Eur. J. Cancer*, vol. 45, no. 2, pp. 228–247, Jan. 2009.
[32] A. Zwanenburg *et al.*, "Assessing robustness of radiomic features by image perturbation," *Scientific Reports*, vol. 9, no. 1. 2019.
[33] I. Shiri *et al.*, "Repeatability of radiomic features in magnetic resonance imaging of glioblastoma: Test-retest and image registration analyses," *Med. Phys.*, vol. 47, no. 9, pp. 4265–4280, Sep. 2020.
[34] W. Lv, S. Ashrafinia, J. Ma, L. Lu, and A. Rahmim, "Multi-Level Multi-Modality Fusion Radiomics: Application to PET and CT Imaging for Prognostication of Head and Neck Cancer," *IEEE J Biomed Health Inform*, vol. 24, no. 8, pp. 2268–2277, Aug. 2020.
[35] S. Ashrafinia, "Quantitative Nuclear Medicine Imaging using Advanced Image Reconstruction and Radiomics," Johns Hopkins University, 2019, 2019.
[36] P. J. Burt and E. H. Adelson, "The Laplacian Pyramid as a Compact Image Code," in *Readings in Computer Vision*, M. A. Fischler and O. Firschein, Eds. San Francisco (CA): Morgan Kaufmann, 1987, pp. 671–679.
[37] A. Toet, "Image fusion by a ratio of low-pass pyramid," *Pattern Recognit. Lett.*, vol. 9, no. 4, pp. 245–253, May 1989.
[38] H. Li, B. S. Manjunath, and S. K. Mitra, "Multisensor Image Fusion Using the Wavelet Transform," *Graphical Models and Image Processing*, vol. 57, no. 3, pp. 235–245, May 1995.
[39] J. J. Lewis, R. J. O'Callaghan, S. G. Nikolov, D. R. Bull, and N. Canagarajah, "Pixel- and region-based image fusion with complex wavelets," *Inf. Fusion*, vol. 8, no. 2, pp. 119–130, Apr. 2007.
[40] F. Nencini, A. Garzelli, S. Baronti, and L. Alparone, "Remote sensing image fusion using the curvelet transform," *Inf. Fusion*, vol. 8, no. 2, pp. 143–156, Apr. 2007.
[41] Q. Zhang and B.-L. Guo, "Multifocus image fusion using the nonsubsampled contourlet transform," *Signal Processing*, vol. 89, no. 7, pp. 1334–1346, Jul. 2009.
[42] G. Piella, "A general framework for multiresolution image fusion: from pixels to regions," *Inf. Fusion*, vol. 4, no. 4, pp. 259–280, Dec. 2003.
[43] J. Zhi-guo, H. Dong-bing, C. Jin, and Z. Xiao-kuan, "A wavelet based algorithm for multi-focus micro-image fusion," in *Third International Conference on Image and Graphics (ICIG'04)*, 2004, pp. 176–179.
[44] Ranjith and Ramesh, "A lifting wavelet transform based algorithm for multi-sensor image fusion," *CRL Tech. J*, 2001.
[45] Hill, Canagarajah, and Bull, "Image Fusion Using Complex Wavelets," *BMVC*, 2002.
[46] Y. Liu, S. Liu, and Z. Wang, "A general framework for image fusion based on multi-scale transform and sparse representation," *Inf. Fusion*, vol. 24, pp. 147–164, Jul. 2015.
[47] B. K. Shreyamsha Kumar, "Image fusion based on pixel significance using cross bilateral filter," *J. VLSI Signal Process. Syst. Signal Image Video Technol.*, vol. 9, no. 5, pp. 1193–1204, Jul. 2015.
[48] G. Pajares and J. Manuel de la Cruz, "A wavelet-based image fusion tutorial," *Pattern Recognit.*, vol. 37, no. 9, pp. 1855–1872, Sep. 2004.
[49] Sahu and Parsai, "Different image fusion techniques–a critical review," *Int. j. mod. educ. comput. sci.*, 2012.